# Robustness of Structured Data Extraction from In-plane Rotated Documents using Multi-Modal Large Language Models (LLM)

[1] Anjanava Biswas , [2] Wrick Talukdar

(1,2) AWS AI & ML, IEEE CIS, California, USA

**Abstract**

Multi-modal large language models (LLMs) have shown remarkable performance in various natural language processing tasks, including data extraction from documents. However, the accuracy of these models can be significantly affected by document *in-plane rotation*, also known as *skew*, a common issue in real-world scenarios for scanned documents. This study investigates the impact of document skew on the data extraction accuracy of three state-of-the-art multi-modal LLMs: Anthropic Claude V3 Sonnet, GPT-4-Turbo, and Llava:v1.6. We focus on extracting specific entities from synthetically generated sample documents with varying degrees of skewness. The results demonstrate that document skew adversely affects the data extraction accuracy of all the tested LLMs, with the severity of the impact varying across models. We identify the safe in-plane rotation angles (SIPRA) for each model and investigate the effects of skew on model hallucinations. Furthermore, we explore existing skew detection and correction mechanisms and discuss their potential limitations. We propose alternative approaches, including developing new multi-modal architectures that are inherently more robust to document skew and incorporating skewing techniques during the pre-training phase of the models. Additionally, we highlight the need for more comprehensive testing on a wider range of document quality and conditions to fully understand the challenges and opportunities associated with using multi-modal LLMs for information extraction in real-world scenarios.

## I. INTRODUCTION

Large Language Models (LLM) are known for their natural language processing (NLP) and natural language understanding (NLU) capabilities; however, these models are often restricted to a single textual modality. This means that traditional LLMs take text as input and generate novel output in textual format. However, the advent of multi-modal LLMs have changed the landscape in terms of processing data in various modalities i.e. text, images, audio and video; and subsequently generate data in different modalities as well. In this paper we examine multi-modal LLM's unique capability of extracting structured data from unstructured or semi-structured documents. These models leverage both text and visual data, allowing for more comprehensive analysis and understanding of document content. However, despite their advanced capabilities, multi-modal LLMs are not immune to common real-world challenges, one of which is document in-plane rotation, or skew.

Skew in documents typically arises during the scanning process, leading to misaligned text that can significantly hinder the accuracy of data extraction tasks. In practical applications, documents such as invoices, forms, and receipts often exhibit varying degrees of skewness, which complicates the extraction of key-value pairs and other critical information. This issue is particularly pertinent in fields such as finance, healthcare, and legal services, where precise data extraction is crucial for operations and decision-making.

Previous research has highlighted the sensitivity of NLP models to various types of noise and distortions in textual data [1,2,3], but the specific impact of document skew on multi-modal LLMs remains underexplored. This study aims to fill this gap by systematically investigating how different degrees of skew affect the performance of these models in extracting key-value pairs from documents. By comparing the extracted data with ground-truth values, we can quantify the degradation in accuracy caused by skew.

Furthermore, this study examines existing mechanisms for skew detection and correction, which are essential for mitigating the adverse effects of skew on data extraction accuracy. By evaluating the performance of these correction methods, we aim to demonstrate their effectiveness in improving the robustness of multi-modal LLMs in real-world document processing scenarios.

**II. PREVIOUS WORK**

The issue of document skew and its impact on data extraction accuracy has been a subject of interest in the field of document analysis and recognition for several decades. Early studies focused on the development of algorithms to detect and correct skew in scanned documents, leveraging techniques such as Hough Transform and projection profiles. These methods laid the groundwork for understanding the significance of skew correction in improving optical character recognition (OCR) accuracy.

In the realm of OCR, Gatos et al. (1997) [4] demonstrated that noise correction on degraded images could significantly enhance the performance of OCR systems by applying a novel locally adaptive thresholding scheme of document binarization, thereby facilitating better character segmentation and recognition. Similarly, Hull (1998) [5] showed that skew detection algorithms based on connected component analysis were effective in realigning skewed documents, leading to improved OCR results.

With the advancements in deep learning techniques, more sophisticated methods for skew detection and correction have been proposed. For instance, Dobai et al. (2019) [6] introduced a neural network-based approach that outperformed traditional methods in terms of accuracy and robustness. Their work highlighted the potential of leveraging machine learning models to address skew detection and skew correction in a more automated and adaptive manner. Document skew correction has been an extensively researched and studied area using techniques such as line detection, nearest-neighbor [7,8] clustering, projection profile [9,10], and axis parallel bounding-boxes [11,12].

Recent advancements in multi-modal LLMs have opened new avenues for document data extraction, but they also bring new challenges related to document presentation. For example, a study by Wu et al. (2023) [13] showed various multi-modal large scale pre-trained model show promising results in different data modalities, for both input and output modalities. There are several other studies which focused on specific multi-modal downstream tasks [14].

It is important to note that our study primarily focuses on entity or key-value pair extractions using multi-modal language models. Entity and key-value pair extraction is a crucial task in document understanding and information retrieval, as it enables the identification of specific pieces of information within a document. Multi-modal LLMs have shown promising results in this domain, leveraging their ability to process and understand both textual and visual information.

One notable work in this area is by Yu et al. (2021), [15] who proposed a method for improving reasoning with contrastive visual information for visual question answering (VQA). Their approach demonstrates the effectiveness of incorporating visual information to enhance the reasoning capabilities of multi-modal LLMs. Although their work focuses on VQA, the principles of leveraging visual information to improve reasoning can be applied to entity and key-value pair extraction tasks as well.

By considering the visual layout and structure of documents, multi-modal LLMs can better understand the relationships between different entities and key-value pairs, leading to more accurate extractions. However, the presence of document skew can disrupt this visual information, potentially impacting the performance of these models. Therefore, investigating the effect of document skew on entity and key-value pair extraction using multi-modal LLMs is a crucial area of research that this study aims to address.

### III.   APPROACH

To investigate the impact of document skew on the data extraction accuracy of multi-modal LLMs, this study will employ a two-part approach. In the first part, we will evaluate the behavior of different multi-modal LLMs, namely GPT-4, Anthropic Claude V3, and Llava-NEXT (also known as Llava:v1.6) [16], when extracting key-value pairs from documents with varying degrees of skewness. The experiment will involve the following steps:

1. Prepare a dataset of documents with ground truth key-value pair annotations.
2. Introduce varying degrees of skew to the documents using image processing techniques.
3. Pass the skewed documents to the selected multi-modal LLMs and extract the key-value pairs.
4. Compare the extracted key-value pairs with the ground truth annotations using evaluation metrics such as precision, recall, and F1-score.
5. Analyze the impact of different degrees of skewness on the extraction accuracy of each multi-modal LLM.

To evaluate the models' robustness to document skew, we generate a series of skewed versions of each document image by applying in-plane rotations in 5-degree increments, starting from 0 degrees up to 355 degrees. This process results in a total of 72 skewed versions for each original document image. We prompt the multi-modal LLMs with each skewed version of the document using a specific JSON schema that defines the structure and format of the expected key-value pairs, as described in the LMDX paper by Perot et al. [17].

By analyzing the average Levenshtein distances across different degrees of skew, we identify the safe in-plane rotation angles (SIPRA) for each model, within which they can maintain relatively high extraction accuracy. We also investigate the critical skew angles beyond which the extraction accuracy deteriorates significantly and the models start producing hallucinations or incorrect values.

In the second part of the study, we explore existing skew detection and correction mechanisms, focusing on the projection profile method. We discuss the potential limitations of these techniques, such as computational overhead and complexity, particularly when dealing with large volumes of documents or noisy and degraded document images.

To address these limitations, we propose alternative approaches for future research and development. One direction is to develop new multi-modal architectures that are inherently more robust to document skew, eliminating the need for explicit de-skewing steps. Another approach is to incorporate skewing techniques during the pre-training phase of the models, exposing them to a diverse range of skewed documents and enabling them to learn robust representations that are invariant to skew.

Furthermore, we highlight the need for more comprehensive testing on a wider range of document quality and conditions, including older, scanned, or stained documents, to fully understand the challenges and opportunities associated with using multi-modal LLMs for information extraction in real-world scenarios.

## IV. METHODOLOGY

This section provides a comprehensive description of the methodology employed in our study to investigate the impact of document skew on the accuracy of key-value pair extraction using multi-modal large language models (LLMs). The methodology involves prompting the models with document images under various degrees of skew, extracting the key-value pairs from the model outputs, and comparing the extracted pairs with the ground truth using the Levenshtein distance metric to evaluate the extraction accuracy.

To establish the ground truth, we manually annotate the key-value pairs in the original, unskewed document images. These ground truth values serve as the reference for evaluating the accuracy of the key-value pairs extracted by the LLMs from the skewed document images.

The Levenshtein distance, also known as the edit distance, is a widely used metric for measuring the dissimilarity between two strings. It calculates the minimum number of single-character edits (insertions, deletions, or substitutions) required to transform one string into another. In the context of our study, we use the Levenshtein distance to quantify the difference between the LLM-extracted key-value pairs and the corresponding ground truth values.

Mathematically, the Levenshtein distance between two strings $a$ and $b$ of lengths $|a|$ and $|b|$, respectively, can be defined using the following recurrence relation:

$$lev_{a,b}(i,j) = \begin{cases} max(i,j) & if\ min(i,j) = 0 \\ min \begin{cases} lev_{a,b}(i, j-1) + 1 \\ lev_{a,b}(i, j-1) + 1 \\ lev_{a,b}(i-1, j-1) + 1_{(a_i \neq b_j)} \end{cases} \end{cases}$$

where $i$ and $j$ are the positions in the strings $a$ and $b$, respectively, and $1_{(a_i \neq b_j)}$ is the indicator function equal to 0 when $a_i = b_j$ and 1 otherwise. The Levenshtein distance can be computed efficiently using dynamic programming. The algorithm fills a matrix $d$ of size $(|a| + 1) \times (|b| + 1)$, where $d[i,j]$ represents the Levenshtein distance between the substrings $a[1..i]$ and $b[1..j]$. The final value $d[|a|, |b|]$ gives the Levenshtein distance between the complete strings $a$ and $b$.

In our study, we calculate the Levenshtein distance between each LLM-extracted key-value pair and its corresponding ground truth value. We then compute the average Levenshtein distance across all key-value pairs for each degree of skew and for each LLM (GPT-4, Claude V3, and Llava). A lower average Levenshtein distance indicates a higher accuracy in key-value pair extraction, while a higher average Levenshtein distance suggests a greater dissimilarity between the extracted pairs and the ground truth. By comparing the average Levenshtein distances for different degrees of skew and across different LLMs, we can assess the impact of document skew on the accuracy of key-value pair extraction and evaluate the robustness of each LLM in handling skewed documents.

**Multi-modal LLM Prompting for Key-Value Pair Extraction**

Multi-modal LLMs, such as GPT-4, Claude V3, and Llava, have the ability to process and understand both image and text inputs. In this study, we leverage this capability to extract key-value pairs from document images by prompting the models with the skewed images of the documents. The prompting process involves providing the models with clear instructions and examples of the desired output format and the types of key-value pairs to be extracted. Following the approach described by Perot et al. [17], we prompt the models with a specific JSON schema that defines the structure and format of the expected key-value pairs. This JSON schema serves as a template for the models to follow when extracting the relevant information from the document images.

The JSON schema used in our study specifies the keys and their corresponding value types for the key-value pairs to be extracted. For example, the schema may include keys such as "firstName" (string), "lastName" (string), "dateOfBirth" (date), and so on. By providing this schema as part of the prompt, we guide the models to focus on extracting the desired key-value pairs and ensure a consistent output format across different document types and layouts. The prompt depicted in the image uses a baseline format; however, for real tests across various multi-modal LLMs, the prompt was converted into a model-compatible format. It is also important to note that while our analysis used single-page image documents, the same methodology applies to multi-page PDFs. In such cases, the PDF pages can first be converted into compatible image types and then fed to the model.

Essentially, we are aiming to perform OCR-like capability with the multi-modal LLMs for structured data extraction. By leveraging the advanced language understanding and visual processing capabilities of these models, we can extract key-value pairs directly from document images without the need for a separate OCR preprocessing step. This approach simplifies the data extraction pipeline and potentially improves the accuracy of the extracted information by taking advantage of the contextual understanding and reasoning abilities of the multi-modal LLMs.

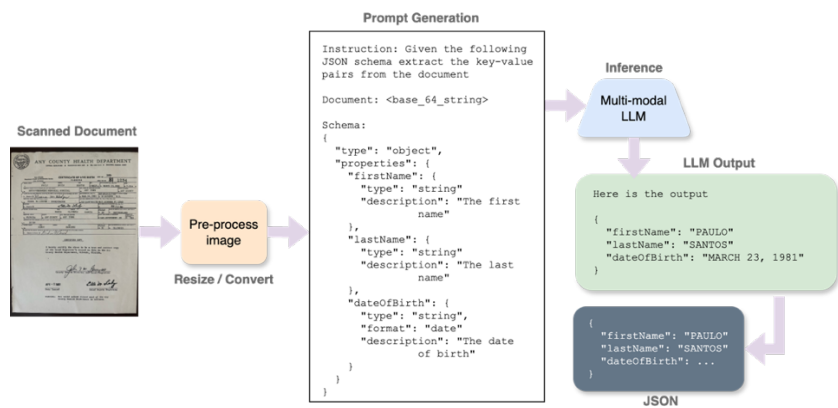

**Figure 1**: Prompting Multi-modal LLM using a derivative of LMDX methodology where the prompt contains the JSON schema of the key-value pairs to be extracted

**Prompting Multi-modal LLMs with Varying Skewness**

In this subsection, we present the methodology and results of prompting the multi-modal LLMs (GPT-4, Claude V3, and Llava) with document images under different degrees of skew. We evaluate the models' performance in extracting key-value pairs accurately and analyze the impact of skewness on the extraction accuracy.

To test the models' robustness to document skew, we generate a series of skewed versions of each document image by applying in-plane rotations in 5-degree increments, starting from 5° up to 355° of in-plane rotation. This process results in a total of 71 skewed versions for each original document image. For each skewed version of the document, we prompt the multi-modal LLMs with the same extraction schema to obtain the key-value pairs. The prompting process follows the approach described in the previous subsection, using a JSON schema to guide the models in extracting the desired information.

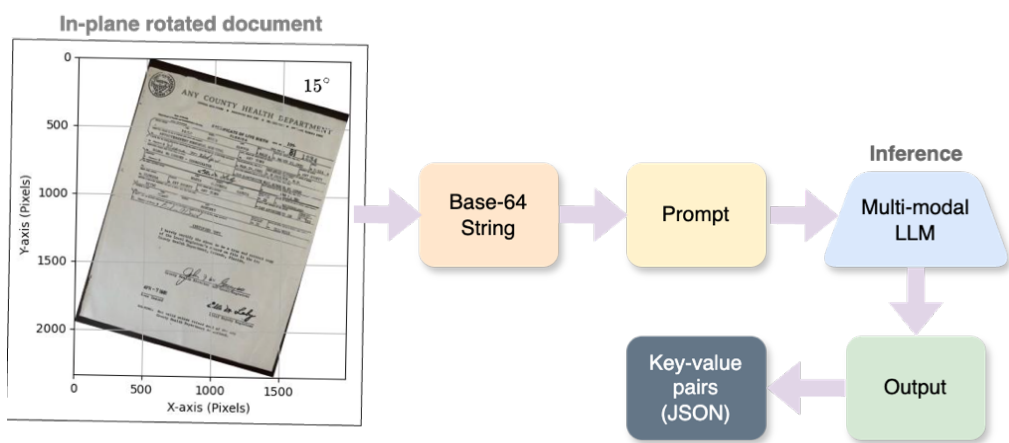

**Figure 2**: Prompting Multi-modal LLM using a derivative of LMDX methodology

where the document is skewed at a given angle

After obtaining the extracted key-value pairs from the LLMs for each skewed document version, we compare them with the ground truth values using the Levenshtein distance metric. The Levenshtein distance, denoted as $lev(a, b)$, measures the minimum number of single-character edits (insertions, deletions, or substitutions) required to transform string $a$ into string $b$.

Mathematically, the Levenshtein distance between two strings $a$ and $b$ of lengths $|a|$ and $|b|$, respectively, can be computed using the following recurrence relation:

$$lev(a,b) = \begin{cases} |a| & if\ |b| = 0 \\ |b| & if\ |a| = 0 \\ lev(tail(a), tail(b)) & if\ a[0] = b[0] \\ 1 + min \begin{cases} lev(tail(a), b) \\ lev(a, tail(b)) \\ lev(tail(a), tail(b)) \end{cases} otherwise \end{cases}$$

where $tail(a)$ and $tail(b)$ represent the substrings of $a$ and $b$ excluding the first character, respectively. We calculate the Levenshtein distance between each extracted key-value pair and its corresponding ground truth value for all 71 skewed versions of each document. This process is repeated for each multi-modal LLM (GPT-4, Claude V3, and Llava) to evaluate their performance under varying degrees of skewness.

V.  RESULTS AND ANALYSIS

To summarize the results, we compute the average Levenshtein distance across all key-value pairs for each degree of skew and for each multi-modal LLM (GPT-4, Claude V3, and Llava). A lower average Levenshtein distance indicates a higher accuracy in key-value pair extraction, while a higher average Levenshtein distance suggests a greater dissimilarity between the extracted pairs and the ground truth.

Consecutively we also analyze the angles of in-plane rotation where the models perform satisfactorily vs. the critical skew angles where the model performance drops significantly. We use a set of two values (entities), namely a person's first name (Value 1) and their last name (Value 2) that are to be extracted from a synthetic sample document by the models. We considered the values as ground truth and applied skew angles in increments of 5 degrees. This forms the basis of the data that we subsequently analyze to determine the impact on Levenshtein distance and critical skew angles.

1. Impact of skew on extraction accuracy

The results demonstrate that document skew has a significant impact on the accuracy of key-value pair extraction for all three multi-modal LLMs tested. As the degree of skew increases, the average Levenshtein distance generally increases, indicating a decrease in extraction accuracy. This trend highlights the importance of considering document skew as a factor in document understanding tasks.

**Anthropic Claude V3 Sonnet**: The graph for Anthropic Claude V3 Sonnet shows that the Levenshtein distance for both the extracted values (Value 1 and Value 2) remains relatively low and stable up to a skew angle of around 25°. Beyond this point, the Levenshtein distance starts to increase sharply, indicating a decline in extraction accuracy. The Levenshtein distance reaches its peak at skew angles near 245° and 335°, suggesting that these extreme skew angles pose the greatest challenge for accurate key-value pair extraction.

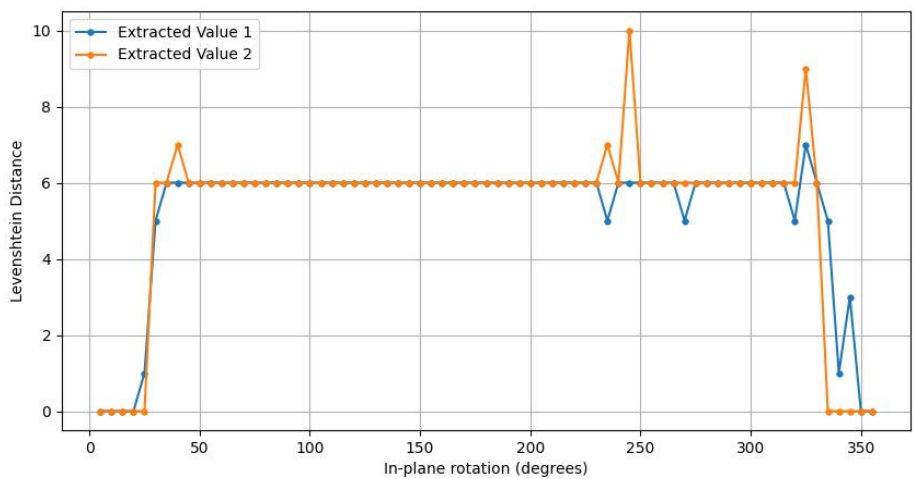

**Figure 3**: Claude V3 Sonnet, Levenshtein distance vs. in-plane rotation angle (skew)

The model exhibits optimal performance on images with skew angles bounded by two intervals: [0°, 25°] and [335°, 360°]. The lower bound of the first interval is 0°, indicating no skew, while the upper bound is 25°, representing a slight clockwise rotation. The second interval's lower bound is 335°, denoting a nearly complete counterclockwise rotation, with the upper bound being 360°, signifying a full rotation back to the starting position. The stabilized Levenshtein distance of 6 are instances where the model was unable to extract both values due to in-plane rotation starting at 45°. These empty values were replaced with a placeholder string <UNKNOWN> for both values.

**OpenAI GPT-4 Turbo**: The graph for OpenAI GPT-4 Turbo exhibits a similar trend to that of Anthropic Claude V3 Sonnet. The Levenshtein distance for the extracted values remains low and stable for skew angles up to approximately 35°. However, beyond this point, the Levenshtein distance starts to increase, indicating a decrease in extraction accuracy. The Levenshtein distances starting at 40° up to 325° vary since the model encountered extreme hallucinations at those angles producing incorrect outputs for both values. The Levenshtein distance subsequently stabilizes to a low 0 starting 330°.

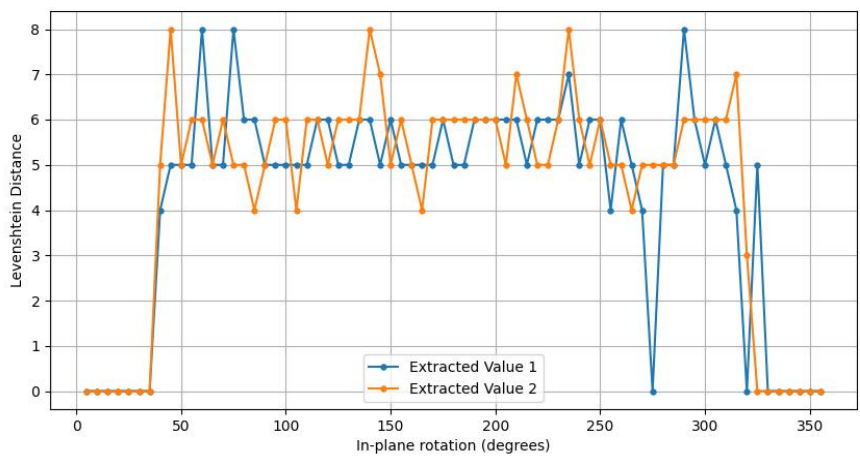

**Figure 4**: GPT-4-turbo, Levenshtein distance vs. in-plane rotation angle (skew)

**Llava v1.6**: The graph for Llava v1.6 presents a slightly different pattern compared to the other two models. The Levenshtein distance for the first name and last name values remains relatively high throughout the entire range of skew angles, with stable values at only a lower range of [5°,10°] and a higher range of [335° to 355°]. However, it is important to note that the average Levenshtein distance of both Value 1 and Value 2 were higher than desirable across the board. There stabilized distances starting 50° up to 315° were due to the fact that the model was unable to extract both values and were subsequently replaced with the placeholder string <UNKNOWN> for both values.

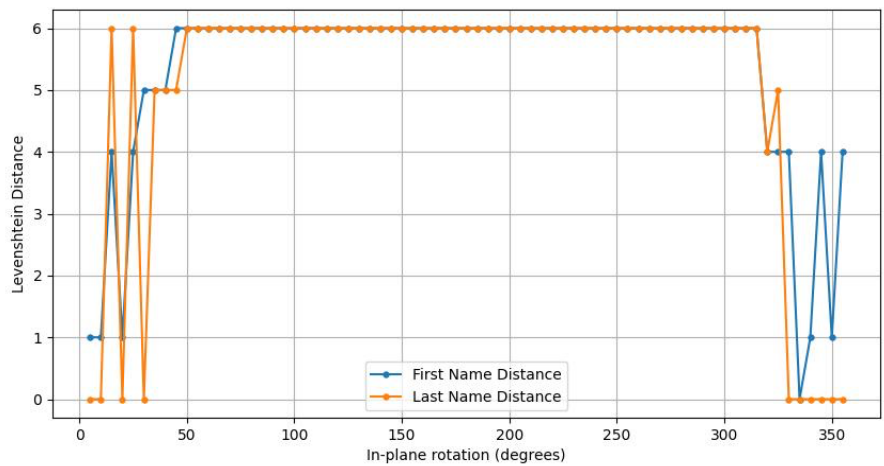

**Figure 5**: Llava:v1.6, Levenshtein distance vs. in-plane rotation angle (skew)

These observations suggest that while all three multi-modal LLMs are affected by document skew to some extent, GPT-4-Turbo exhibits a higher degree of resilience to skew-induced extraction errors. The critical skew angles, particularly from the range of ~30° up to ~335°, pose significant challenges for Anthropic Claude V3 Sonnet and OpenAI GPT-4 Turbo, leading to a notable decrease in extraction accuracy. On the other hand, Llava v1.6 has a relatively lesser tolerance of skew angles with ranges [0°, 10°] or [335°, 355°] being the only tolerable ranges for somewhat successful extraction. Note that we use pre-trained Llava:v1.6 for our test and no fine-tuning was performed.

### 2. Critical skew angles

The analysis reveals critical skew angles at which the extraction accuracy starts to deteriorate significantly. For example, we may observe that the average Levenshtein distance remains relatively low for skew angles up to 30 degrees but begins to increase rapidly beyond that point. Identifying these critical skew angles is crucial for determining the acceptable range of skewness for reliable key-value pair extraction.

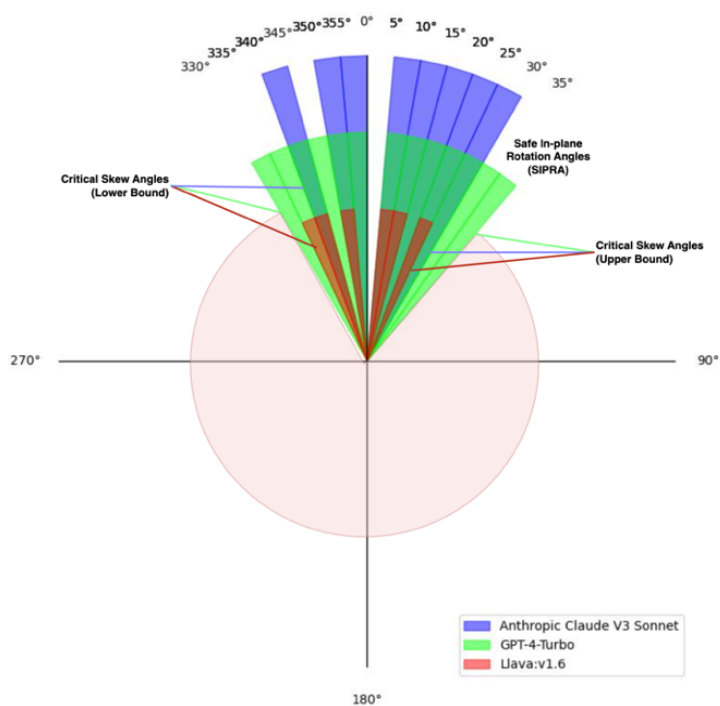

**Figure 6**: Safe in-plane rotation angles (SIPRA) for Sonnet, GPT-4-turbo, and Llava:v1.6

The diagram provided illustrates the concept of critical skew angles and safe in-plane rotation angles (SIPRA) for the three multi-modal LLMs: Anthropic Claude V3 Sonnet, GPT-4-Turbo, and Llava:v1.6. The circular chart is divided into different sectors, each representing a specific range of skew angles for the respective models.

The green sectors represent the SIPRA for GPT-4-turbo, which are the ranges of skew angles where the models can perform key-value pair extraction with relatively high accuracy. These ranges are typically centered around 0 degrees (no skew) and extend symmetrically in both clockwise and counterclockwise directions.

The red sectors, on the other hand, represent the SIPRA for Llava:v1.6, which are the ranges of skew angles where the model can perform potentially satisfactorily. Similarly, the purple sectors represents the SIPRA for Anthropic Claude V3 Sonnet model.

Based on the diagram, we can infer the following:

1. GPT-4-Turbo (green sector) has the widest range of safe in-plane rotation angles. This suggests that GPT-4-Turbo is the most robust among the three models in terms of handling document skew and maintaining high extraction accuracy within this range.
2. Anthropic Claude V3 Sonnet (blue sector) has a slightly narrower range of safe in-plane rotation angles compared to GPT-4-Turbo. While the model can handle some degree of document skew, its performance may start to degrade more quickly beyond this range compared to GPT-4-Turbo.
3. Llava:v1.6 (red sector) has the narrowest range of safe in-plane rotation angles among the three models. This indicates that Llava (pre-trained only) is the most sensitive to document skew, and its extraction accuracy may start to deteriorate at smaller skew angles compared to the other two models.

**3. Effects skew on model hallucinations**

In addition to the impact on extraction accuracy, document skew also influences the tendency of multi-modal LLMs to generate hallucinations, i.e., producing incorrect or fabricated information. The behavior of the three models—Anthropic Claude V3 Sonnet, GPT-4-Turbo, and Llava:v1.6—varies when the skew angles exceed their respective safe in-plane rotation angles (SIPRA).

1. Anthropic Claude V3 Sonnet: Claude V3 Sonnet demonstrated a lower propensity for hallucinations beyond its SIPRA zone (the purple zone). When the skew angles exceeded the critical angles, Claude V3 Sonnet adhered to the provided instruction in its system prompt and returned no values. This behavior suggests that Claude V3 Sonnet has a built-in mechanism to handle cases where the extraction accuracy is likely to be poor, opting to return no information rather than generating potentially incorrect or fabricated values. This conservative approach taken by Claude V3 Sonnet can be beneficial in scenarios where the reliability and trustworthiness of the extracted information are of utmost importance. By returning no values beyond the critical skew angles, Claude V3 Sonnet minimizes the risk of propagating erroneous or misleading information downstream in the data processing pipeline.

2. GPT-4-Turbo: GPT-4-Turbo exhibited the highest tendency for hallucinations among the three models when the skew angles exceeded its SIPRA zone (the green zone). For skew angles beyond the critical angles, GPT-4-Turbo produced almost all incorrect values, indicating a significant degree of hallucination, despite having explicitly instructed in the system prompt. This behavior suggests that GPT-4-Turbo may not have a robust mechanism to handle cases where the extraction accuracy is compromised due to excessive document skew. Instead of returning no values or indicating low confidence, GPT-4-Turbo generates incorrect information, potentially leading to the propagation of errors and inconsistencies in the extracted data. The high degree of hallucination exhibited by GPT-4-Turbo beyond its SIPRA zone highlights the importance of ensuring that the document skew is within the safe range before applying this model for key-value pair extraction. Failure to do so may result in the generation of unreliable and misleading information.

3. Llava:v1.6: Llava:v1.6 demonstrated behavior similar to Claude V3 Sonnet in terms of hallucinations beyond its SIPRA zone (the red zone). When the skew angles exceeded the critical angles, Llava:v1.6 also returned no values, indicating a conservative approach to handling cases where the extraction accuracy is likely to be poor. Like Claude V3 Sonnet, Llava:v1.6's behavior suggests the presence of a mechanism to prevent the generation of incorrect or fabricated information when the document skew is beyond the acceptable range. By returning no values, Llava:v1.6 prioritizes the reliability and trustworthiness of the extracted information, even if it means providing no information at all.

The varying degrees of hallucination exhibited by the three models beyond their respective SIPRA zones underscore the importance of considering not only the extraction accuracy but also the model's behavior when dealing with excessive document skew. While GPT-4-Turbo may have a wider SIPRA zone, its high tendency for hallucinations beyond the critical angles raises concerns about the reliability of the extracted information.

On the other hand, the conservative approach taken by Claude V3 Sonnet and Llava:v1.6, where they return no values beyond the critical skew angles, helps maintain the integrity of the extracted data by minimizing the risk of propagating incorrect or fabricated information.

When selecting a multi-modal LLM for key-value pair extraction from skewed documents, it is crucial to consider both the model's SIPRA zone and its behavior beyond the critical angles. The choice of model should align with the specific requirements of the application, balancing the need for a wider SIPRA zone with the importance of minimizing hallucinations and ensuring the reliability of the extracted information. Bai et al. (2024) [18] has extensively studied several reasons a multi-modal LLM may hallucinate over vision, audio, and other modalities and have studied how several mechanisms of mitigating hallucinations such as LRV-Instruction, HalluciDoctor, ReCaption, and EOS Decision [19,20,21,22]. These mitigation strategies may be helpful in tackling model hallucination; however, mitigation may begin with document pre-conditioning by identifying skew in a document and taking steps to de-skew, discussed in the following section.

## VI. ADDRESSING IN-PLANE ROTATION

The results of our study highlight the significant impact of document skew on the extraction accuracy of multi-modal LLMs. As demonstrated by the analysis of critical skew angles and the safe in-plane rotation angles (SIPRA) for Anthropic Claude V3 Sonnet, GPT-4-Turbo, and Llava:v1.6, the performance of these models deteriorates when the document skew exceeds certain thresholds. Therefore, it is crucial to address the issue of document skew to ensure high extraction accuracy and reliable results.

One approach to mitigate the impact of document skew is to apply de-skewing techniques before feeding the documents into the multi-modal LLMs. De-skewing involves estimating the skew angle of the document and applying a correction to align the text lines horizontally. Several common de-skewing techniques have been proposed in the literature, including:

1. Line detection: This method involves identifying the text lines in the document image and estimating the skew angle based on the orientation of these lines. Techniques such as Hough transform or connected component analysis can be used to detect the lines and calculate the skew angle.

2. Nearest-neighbor clustering: This approach relies on grouping together nearby pixels or connected components in the document image and estimating the skew angle based on the orientation of these clusters. The assumption is that the text lines will form distinct clusters that can be used to determine the overall skew of the document.

3. Projection profile: This technique involves projecting the document image onto a vertical or horizontal axis and analyzing the resulting profile. The skew angle is estimated by finding the angle that maximizes the variation in the projection profile, indicating the alignment of the text lines.

4. Axis-parallel bounding boxes: This method involves fitting axis-parallel bounding boxes around the connected components in the document image and estimating the skew angle based on the orientation of these bounding boxes. The assumption is that the majority of the bounding boxes will be aligned with the text lines, providing an estimate of the overall skew.

While these de-skewing techniques have been shown to be effective in improving the extraction accuracy of multi-modal LLMs, they may also introduce additional computational overhead and complexity to the document processing pipeline. Applying de-skewing algorithms to large volumes of documents can be time-consuming and resource-intensive, potentially limiting the scalability and efficiency of the extraction process.

Moreover, the performance of these de-skewing techniques may vary depending on the specific characteristics of the documents, such as the presence of graphics, tables, or handwritten text. In some cases, the de-skewing algorithms may struggle to accurately estimate the skew angle, leading to suboptimal correction and potentially impacting the extraction accuracy.

Given these limitations, it is worth exploring alternative approaches to address the issue of document skew in multi-modal LLMs. One potential direction is the development of new multi-modal architectures that are inherently more robust to document skew. By designing models that can effectively handle skewed documents without the need for explicit de-skewing, we can simplify the document processing pipeline and improve the efficiency of the extraction process.

Another approach is to incorporate skewing techniques during the pre-training phase of the multi-modal LLMs. By exposing the models to a diverse range of skewed documents during training and teaching them to learn robust representations that are invariant to skew, we can potentially enhance their ability to handle skewed documents at inference time. This approach would eliminate the need for a separate de-skewing step and enable the models to directly process skewed documents with improved accuracy.

The idea behind this approach is to simulate the presence of skewed documents during the pre-training phase, allowing the models to learn features and representations that are resilient to the effects of skew. By training the models on a large corpus of artificially skewed documents, along with their corresponding ground truth information, the models can learn to extract relevant features and information regardless of the level of skew present in the input.

This approach leverages the power of transfer learning, where the knowledge gained during pre-training on skewed documents can be transferred to the downstream task of information extraction from real-world skewed documents. By incorporating skewing techniques during pre-training, we can potentially improve the generalization ability of the multi-modal LLMs and enhance their performance on skewed documents without the need for explicit de-skewing.

### VII.   LIMITATIONS OF OUR EXPERIMENT

While our study provides valuable insights into the impact of document skew on the extraction accuracy of multi-modal LLMs, it is important to acknowledge the limitations of our experimental setup. One significant limitation is that our experiments primarily involved dealing with mostly pristine documents, which may not fully represent the diverse range of document quality encountered in real-world scenarios.

In practice, documents often come in various conditions, including older, scanned, or stained documents that introduce additional noise and artifacts. These factors can further complicate the extraction process and exacerbate the impact of document skew on the performance of multi-modal LLMs.

During our study, we noticed that when dealing with older, scanned, or stained documents, the performance of the multi-modal LLMs deteriorated even more rapidly compared to pristine documents. The presence of noise, such as uneven illumination, background clutter, or physical damage to the document, can make it more challenging for the models to accurately identify and extract the relevant information.

As a result of this increased noise, we observed that the safe in-plane rotation angles (SIPRA) zones for the multi-modal LLMs became smaller when dealing with these types of documents. The critical skew angles, beyond which the extraction accuracy significantly declines, were reached earlier compared to pristine documents. This suggests that the models are less tolerant to skew when dealing with noisy or degraded documents.

Moreover, we noticed an increased tendency for the multi-modal LLMs to generate hallucinations or incorrect information when processing older, scanned, or stained documents. The additional noise and artifacts present in these documents can confuse the models and lead to more frequent generation of irrelevant or fabricated information.

These observations highlight the need for more comprehensive testing and evaluation of multi-modal LLMs on a wider range of document quality and conditions. While our study provides a foundation for understanding the impact of document skew, it is crucial to expand the scope of future experiments to include a diverse set of real-world documents, encompassing various levels of noise, degradation, and complexity.

By conducting more extensive tests on a broader range of document types and conditions, we can gain a more comprehensive understanding of the limitations and challenges associated with using multi-modal LLMs for information extraction. This will enable us to develop more robust and adaptable models that can handle the variability and noise present in real-world document processing tasks.

Furthermore, our findings underscore the importance of pre-processing techniques, such as noise reduction and image enhancement, in addition to de-skewing, when dealing with older, scanned, or stained documents. By applying appropriate pre-processing steps to mitigate the impact of noise and artifacts, we can potentially improve the extraction accuracy and reduce the occurrence of hallucinations in multi-modal LLMs.

## VIII.   CONCLUSION

In this study, we investigated the impact of document skew on the accuracy of key-value pair extraction using multi-modal large language models (LLMs). We focused on three state-of-the-art models: Anthropic Claude V3 Sonnet, GPT-4-Turbo, and Llava:v1.6. Our experimental setup involved extracting specific entities, namely a person's first name and last name, from synthetically generated sample documents. We introduced varying degrees of skew to these documents in increments of 5 degrees, ranging from 0 to 355 degrees, and evaluated the extraction accuracy using the Levenshtein distance metric.

Our analysis revealed several important findings. First, we observed that document skew has a significant impact on the extraction accuracy of all three multi-modal LLMs tested. As the degree of skew increased, the average Levenshtein distance between the extracted values and the ground truth values generally increased, indicating a decrease in extraction accuracy. This highlights the importance of considering document skew as a critical factor in document understanding tasks.

Second, we identified the safe in-plane rotation angles (SIPRA) for each model, which represent the range of skew angles within which the models can maintain relatively high extraction accuracy. GPT-4-Turbo demonstrated the widest SIPRA zone, followed by Llava:v1.6 and Anthropic Claude V3 Sonnet. Beyond these SIPRA zones, the extraction accuracy deteriorated significantly, indicating the presence of critical skew angles that pose challenges for accurate key-value pair extraction.

Third, we investigated the effects of skew on model hallucinations, i.e., the generation of incorrect or fabricated information. We found that Anthropic Claude V3 Sonnet and Llava:v1.6 exhibited a lower propensity for hallucinations beyond their SIPRA zones, often returning no values as per their system prompts. In contrast, GPT-4-Turbo demonstrated a higher tendency for hallucinations, producing mostly incorrect values beyond its critical skew angles.

Our findings underscore the importance of addressing document skew to ensure high extraction accuracy and reliable results when using multi-modal LLMs. We discussed the potential of applying de-skewing techniques, such as line detection, nearest-neighbor clustering, projection profile, and axis-parallel bounding boxes, as a pre-processing step to mitigate the impact of skew. However, we also acknowledged the computational overhead and complexity associated with these techniques, particularly when dealing with large volumes of documents.

To overcome these limitations, we proposed exploring alternative approaches, such as developing new multi-modal architectures that are inherently more robust to document skew and incorporating skewing techniques during the pre-training phase of the models. By exposing the models to a diverse range of skewed documents during training and enabling them to learn robust representations, we can potentially enhance their ability to handle skewed documents at inference time without the need for explicit de-skewing. Furthermore, we recognized the limitations of our experimental setup, which primarily involved dealing with pristine documents. We observed that the performance of multi-modal LLMs deteriorated even more rapidly when dealing with older, scanned, or stained documents, resulting in smaller SIPRA zones and increased hallucinations. This highlights the need for more comprehensive testing on a wider range of document quality and conditions to fully understand the challenges and opportunities associated with using multi-modal LLMs for information extraction in real-world scenarios.

In conclusion, our study provides valuable insights into the impact of document skew on the accuracy of key-value pair extraction using multi-modal LLMs. We have shown that document skew significantly affects the performance of these models, and we have identified the SIPRA zones and critical skew angles for three state-of-the-art models. Our findings emphasize the importance of addressing document skew, either through de-skewing techniques or by developing more robust multi-modal architectures and training approaches.

However, our study also highlights the need for further research and experimentation to fully understand the limitations and challenges associated with using multi-modal LLMs for information extraction in real-world scenarios. Future work should focus on expanding the scope of experiments to include a wider range of document quality and conditions, exploring alternative approaches to handling document skew, and developing more comprehensive evaluation metrics and benchmarks. By advancing our understanding of the impact of document skew on multi-modal LLMs and developing more robust and efficient approaches to information extraction, we can unlock the full potential of these powerful models in real-world document understanding tasks. This will enable organizations to extract valuable insights from large volumes of unstructured documents, streamline their processes, and make more informed decisions based on accurate and reliable data.

## IX. REFERENCES

[1] Kartikay Bagla, Shivam Gupta, Ankit Kumar, and Anuj Gupta. 2024. Noisy Text Data: foible of popular Transformer based NLP models. In Proceedings of the Third International Conference on AI-ML Systems (AIMLSystems '23). Association for Computing Machinery, New York, NY, USA, Article 33, 1–5. https://doi.org/10.1145/3639856.3639889

[2] Zhang, Yunxiang, Liangming Pan, Samson Tan, and Min-Yen Kan. "Interpreting the robustness of neural NLP models to textual perturbations." arXiv preprint arXiv:2110.07159 (2021).

[3] Bagla, Kartikay, Ankit Kumar, Shivam Gupta, and Anuj Gupta. "Noisy Text Data: Achilles' Heel of popular transformer based NLP models." arXiv preprint arXiv:2110.03353 (2021).

[4] Gatos, B., Pratikakis, I., & Perantonis, S. J. (1997). Adaptive degraded document image binarization. Pattern Recognition, 39(3), 317-327.

[5] Hull, Jonathan. (1998). Document image skew detection: Survey and annotated bibliography. 10.1142/9789812797704_0003.

[6]   Dobai, Lorand, and Mihai Teletin. "A document detection technique using convolutional neural networks for optical character recognition systems." In ESANN. 2019.

[7]   Aradhya V., Kumar G.H., Shivakumara P. An accurate and efficient skew estimation technique for South Indian documents: A new boundary growing and nearest neighbor clustering based approach. Int. J. Robot. Autom. 2007;22:272–280. doi: 10.2316/Journal.206.2007.4.206-2992.

[8]   Al-Khatatneh A., Pitchay S.A., Al-Qudah M. A Review of Skew Detection Techniques for Document; Proceedings of the 17th UKSIM-AMSS International Conference on Modelling and Simulation; Washington, DC, USA. 25–27 March 2015; pp. 316–321.

[9]   Sun S.J. Skew detection using wavelet decomposition and projection profile analysis. Pattern Recognit. Lett. 2007;28:555–562.

[10]  Bekir Y. Projection profile analysis for skew angle estimation of woven fabric images. J. Text. Inst. Part 3 Technol. New Century.

[11]  Shafii M. Optical Character Recognition of Printed Persian/Arabic Documents. University of Windsor (Canada); Windsor, ON, Canada: 2014.

[12]  Mascaro A.A., Cavalcanti R.D.C., Mello R.A.B. Fast and robust skew estimation of scanned documents through background area information. Pattern Recognit. Lett. 2010;31:1403–1411. doi: 10.1016/j.patrec.2010.03.016.

[13]  Wu, Jiayang, Wensheng Gan, Zefeng Chen, Shicheng Wan, and S. Yu Philip. "Multimodal large language models: A survey." In 2023 IEEE International Conference on Big Data (BigData), pp. 2247-2256. IEEE, 2023.

[14]  Summaira, Jabeen, Xi Li, Amin Muhammad Shoib, and Jabbar Abdul. "A review on methods and applications in multimodal deep learning." arXiv preprint arXiv:2202.09195 (2022).

[15]  Yu Long, Pengjie Tang, Hanli Wang, and Jian Yu. 2021. Improving reasoning with contrastive visual information for visual question answering. Electronics Letters 57, 20 (2021), 758–760

[16]  Haotian Liu, Chunyuan Li, Yuheng Li, Bo Li, Yuanhan Zhang, Sheng Shen, and Yong Jae Lee. 2024b. LLaVA-NeXT: Improved reasoning, OCR, and world knowledge.

[17]  Perot, Vincent, Kai Kang, Florian Luisier, Guolong Su, Xiaoyu Sun, Ramya Sree Boppana, Zilong Wang, Jiaqi Mu, Hao Zhang, and Nan Hua. "LMDX: Language Model-based Document Information Extraction and Localization." arXiv preprint arXiv:2309.10952 (2023).

[18]  Bai, Zechen, Pichao Wang, Tianjun Xiao, Tong He, Zongbo Han, Zheng Zhang, and Mike Zheng Shou. "Hallucination of Multimodal Large Language Models: A Survey." *arXiv preprint arXiv:2404.18930* (2024).

[19]  Liu, Fuxiao, Kevin Lin, Linjie Li, Jianfeng Wang, Yaser Yacoob, and Lijuan Wang. "Aligning large multi-modal model with robust instruction tuning." *arXiv preprint arXiv:2306.14565* (2023).

[20] Yu, Qifan, Juncheng Li, Longhui Wei, Liang Pang, Wentao Ye, Bosheng Qin, Siliang Tang, Qi Tian, and Yueting Zhuang. "Hallucidoctor: Mitigating hallucinatory toxicity in visual instruction data." *arXiv preprint arXiv:2311.13614* (2023).

[21] Lei Wang, Jiabang He, Shenshen Li, Ning Liu, and Ee-Peng Lim. 2023. Mitigating Fine-Grained Hallucination by Fine-Tuning Large Vision-Language Models with Caption Rewrites. arXiv preprint arXiv:2312.01701 (2023).

[22] Zihao Yue, Liang Zhang, and Qin Jin. 2024. Less is More: Mitigating Multimodal Hallucination from an EOS Decision Perspective. arXiv preprint arXiv:2402.14545 (2024).